%% file: main.tex
\documentclass[letterpaper, 10 pt, journal, twoside]{IEEEtran}
\usepackage{bm}
\usepackage{graphicx}
\usepackage{comment}
\usepackage{amsmath,amssymb} 
\usepackage{color}
\usepackage{subfig}
\usepackage{mathtools}
\usepackage{xfrac}
\usepackage{booktabs}
\usepackage{multirow}
\usepackage{nicefrac}
\usepackage{booktabs}
\usepackage{caption}
\usepackage{hyperref}

\ifCLASSINFOpdf
\else
\fi
\begin{document}
%
\title{Learning Optical Flow, Depth, and Scene Flow without Real-World Labels}
%
%
%

\author{Vitor Guizilini \quad Kuan-Hui Lee \quad Rareș Ambruș \quad Adrien Gaidon
\vspace{0.2mm}
\\
Toyota Research Institute (TRI), Los Altos, CA 
\\
{\tt\small \{first.lastname\}@tri.global}

%
}
%
%

\markboth{}
{} 

%



\maketitle

\begin{abstract}
\input{sections/00abstract}

\end{abstract}


\input{sections/01introduction}
\input{sections/02related}

\input{sections/03methodology}

\input{sections/04experiments}

\input{sections/05conclusion}

%
\IEEEpeerreviewmaketitle



\ifCLASSOPTIONcaptionsoff
  \newpage
\fi



%



\bibliographystyle{IEEEtran}
\bibliography{references}

%








\end{document}

%% file: sections/00abstract.tex
Self-supervised monocular depth estimation enables robots to learn 3D perception from raw video streams. This scalable approach leverages projective geometry and ego-motion to learn via view synthesis, assuming the world is mostly static.
Dynamic scenes, which are common in autonomous driving and human-robot interaction, violate this assumption. Therefore, they require modeling dynamic objects explicitly, for instance via estimating pixel-wise 3D motion, i.e. scene flow.
However, the simultaneous self-supervised learning of depth and scene flow is ill-posed, as there are infinitely many combinations that result in the same 3D point. 
In this paper we propose DRAFT, a new method capable of jointly learning depth, optical flow, and scene flow by combining synthetic data with geometric self-supervision.
Building upon the RAFT architecture, we learn optical flow as an intermediate task to bootstrap depth and scene flow learning via triangulation.
Our algorithm also leverages temporal and geometric consistency losses across tasks to improve multi-task learning.
Our DRAFT architecture simultaneously establishes a new state of the art in all three tasks in the self-supervised monocular setting on the standard KITTI benchmark. Project page: \url{https://sites.google.com/view/tri-draft}.

%% file: sections/01introduction.tex
\section{Introduction}

\IEEEPARstart{S}{elf-supervised} learning on videos for 3D perception replaces explicit supervision by geometric priors.
Using geometric view synthesis as a learning objective, this approach has been successfully applied to a wide range of key robot vision tasks in the challenging monocular setting, including the estimation of \emph{ego-motion}~\cite{packnet-twostream,yang2018deep}, the 6 degree-of-freedom camera translation and rotation; \emph{depth}~\cite{packnet,monodepth2,godard2017unsupervised,manydepth}, the per-pixel distance value from the image plane; \textit{optical flow}~\cite{hd3,pwcnet,flownet,flownet2,vcn}, the 2D pixel displacement between frames; and \emph{scene flow}~\cite{epc,epc++,selfsceneflow,multi-scene-flow}, the 3D motion of each point in the scene.
%
%
%
Although these tasks are clearly related~\cite{epc,epc++, deep_mv_depth}, they typically require stereo pairs at training time~\cite{selfsceneflow,multi-scene-flow} to resolve reprojection ambiguities in self-supervision. 

In this paper, we instead propose an architecture capable of \emph{jointly} learning optical flow, depth, and scene flow on raw videos by leveraging additional synthetic data. We build upon the recent RAFT architecture~\cite{raft} for optical flow and extend it to jointly estimate depth and scene flow in a multi-task framework.  Our DRAFT (Depth-RAFT) architecture combines synthetic supervision and real-world unlabeled data in a mixed-batch training schedule to simultaneously resolve geometric ambiguities and facilitate sim-to-real domain transfer. Importantly, this enables the joint learning of depth and scene flow without explicit depth supervision~\cite{raft3d} or stereo pairs~\cite{epc,epc++,selfsceneflow,multi-scene-flow} at training time. 

%
%
In summary, the \textbf{key contributions} of our work are: 
\begin{itemize}
\item A novel architecture for the \textbf{multi-task learning of depth, optical flow, and scene flow in a monocular self-supervised setting}, leveraging synthetic data for domain transfer. To the best of our knowledge this is the first self-supervised monocular architecture that \textbf{does not use stereo at training time}.
\textcolor{black}{
\item A study showing how different techniques improve performance in this novel setting, including (1)
\textbf{triangulated depth from optical flow as initialization} for depth and scene flow; (2) \textbf{geometric consistency losses} to ensure task alignment; and (3) computing \textbf{efficient forward-backward estimates} for temporal consistency.}
\item We report \textbf{state of the art results} on the KITTI dataset in all three considered tasks, \textbf{using the same model}. 
\end{itemize}

%% file: sections/02related.tex
\section{Related Work}

\input{figures/diagram}

\subsection{Self-Supervised Depth Estimation}

Early approaches to learning-based depth estimation were supervised~\cite{eigen2014depth}.
Follow-up works introduced self-supervision to monocular depth estimation, both in the stereo~\cite{godard2017unsupervised} and monocular~\cite{zhou2017unsupervised} settings. 
These works use geometric constraints to warp color information from one viewpoint to another, minimizing the reprojection error and in the process learning depth as a proxy task. 
Learning from raw videos massively increases scalability, leading to self-supervised results comparable to or even surpassing methods trained with ground-truth supervision~\cite{packnet}. 
\textcolor{black}{Recent works further extend self-supervised monocular depth estimation to surface normal calculation~\cite{surfacenormals}, the use of semantic information~\cite{packnet-semguided}, additional losses~\cite{monodepth2}, improvements in pose estimation~\cite{packnet-twostream,zhao2020towards}, unknown camera geometries \cite{gordon2019depth,vasiljevic2020neural}, multi-task domain adaptation \cite{guda}, and multi-frame inference~\cite{manydepth}.}

Two key limitations in self-supervised monocular depth estimation are: (1) the generation of \textit{scale-ambiguous} models, and (2) a \textit{static world} assumption. 
Scale can be recovered thanks to weak supervision (e.g., velocity estimates in~\cite{packnet} or camera extrinsics in~\cite{guizilini2021surround}).
To address the static world assumption, most methods propose loss masks~\cite{monodepth2,manydepth}, or alternatively consider piece-wise rigid motion~\cite{sfmnet}. 
In contrast, our architecture addresses these limitations by (1) using synthetic data with a similar camera geometry to real-world data, to facilitate scale transfer between domains; and (2) jointly generating depth and scene flow estimates, thus explicitly modeling dynamic objects.  

\subsection{Optical Flow Estimation}

Optical flow, finding the 2D pixel displacement field between two images, is a core computer vision problem~\cite{opticalflow_survey}.
%
%
FlowNet~\cite{flownet,flownet2}  introduced deep learning to this task, proposing an encoder-decoder architecture to directly regress dense optical flow from two frames, using synthetic data for pretraining as a source of geometric priors~\cite{sintel,Mayer2016ALD}.
A key component of these approaches is coarse-to-fine processing~\cite{hd3,Hur:2019:IRR,vcn}, in which multiple resolutions are used to generate and refine flow fields.
%

Other works focus instead on iterative refinement for optical flow and related tasks.
SpyNet~\cite{SpyNet}, PWC-Net~\cite{pwcnet} and VCN~\cite{vcn} apply iterative refinement using coarse-to-fine pyramids.
Some works propose to use self-supervised learning methods with photometric loss \cite{jason2016back,SelFlow} or forward-backward consistency \cite{sundaram2010dense,SelFlow}.
Recently, RAFT \cite{raft} was proposed as a recurrent neural network architecture for optical flow estimation. 
A 4D correlation volume is constructed by computing the visual similarity between pixel pairs, and during inference a recurrent update operator indexes from this correlation volume to update a high-resolution flow field.
Our proposed DRAFT (Depth-RAFT) architecture builds upon RAFT, extending it to (1) a multi-task setting where depth and scene flow estimates are jointly generated in addition to optical flow; and (2) a self-supervised monocular setting, by eliminating the need for real-world labels or stereo pairs during training.


\subsection{Scene Flow Estimation}

The concept of scene flow was first introduced in \cite{sceneflow}, consisting of dense 3D motion estimation for each point in the scene.
Earlier works focused on variational formulations or graphical models, yielding limited accuracy \cite{varsceneflow} and/or slow runtimes \cite{runtimesceneflow,objsceneflow}. 
Recently, deep learning methods \cite{selfsceneflow,Ilg_2018_ECCV,raft3d,saxena2019pwoc} have achieved state-of-the-art performance in real-time with a combination of large-scale pre-training on synthetic datasets followed by fine-tuning on the target
domain with a smaller subset of annotated samples.
To facilitate the joint learning of depth and scene flow, stereo-based methods~\cite{effiscene} use a disparity map to recover the scene structure in 3D space as well as dense scene flow from adjacent frames.
Other methods use sequences of RGBD images~\cite{piece-wise-scnflow,Lv18eccv} or range-based pointclouds~\cite{raft3d,sceneflow3d,wu2020pointpwc,pvraft} as a source of sparse 3D information.

Self-supervised scene flow methods~\cite{epc,selfsceneflow,multi-scene-flow,ijcai2019-123} have also been developed to alleviate the need for ground-truth real-world data.
Multi-task networks for the joint learning of optical flow and depth have been proposed~\cite{epc,epc++,chen2019self,yin2018geonet}, where scene flow can be indirectly recovered. However, these methods cannot reason over occluded pixels, since it can only recover areas with valid reprojection between frames. %
Our method extends this setting to include explicit scene flow estimation, and a series of cross-task consistency losses between these two sources of scene flow.
To the best of our knowledge, ours is the first self-supervised monocular scene flow method that does not use stereo at training time~\cite{epc,epc++,selfsceneflow,multi-scene-flow}, focusing instead on single cameras in both synthetic and real-world domains.


%% file: figures/diagram.tex

\begin{figure*}[t!]
    \centering
    \includegraphics[width=0.9\textwidth]{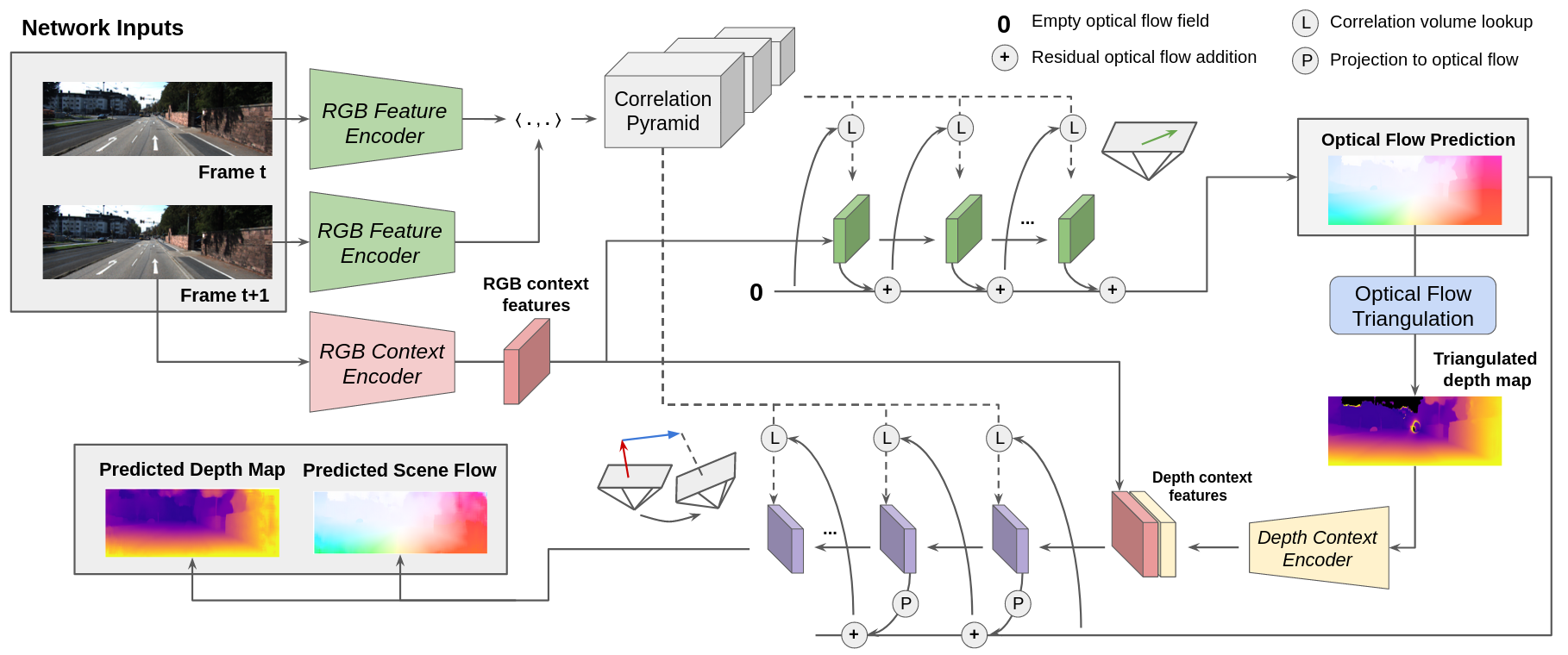}
    \caption{\textbf{DRAFT}, our proposed multi-task architecture for jointly learning optical flow, depth, and scene flow from videos.}
    \label{fig:diagram2}
\vspace{-3mm}
\end{figure*}

%% file: sections/03methodology.tex
\section{The DRAFT Architecture}
\label{sec:architecture}

\input{figures/optflow_depth}

A diagram depicting our proposed architecture is shown in Figure \ref{fig:diagram2}. It is composed of two stages: optical flow estimation (top), followed by depth and scene flow estimation (bottom). These two stages use the same input information: two RGB images $I_t$ and $I_{t+1}$ of resolution $H \times W$ with known intrinsics $\mathbf{K}$ and relative pose
$ \mathbf{T}_t^{t+1} = 
\big|\begin{smallmatrix}
  \mathbf{R} & \mathbf{t} \\
  \mathbf{0} & 1
\end{smallmatrix}\big|$.
Additionally, depth and scene flow estimates use optical flow predictions as additional input, both via optical flow triangulation to produce initial depth estimates, and as initialization to query the correlation pyramid. Formally, we aim to recover the function $f:(I_t,I_{t+1}) \to (\textbf{d},\textbf{o},\textbf{s})$ that recovers depth $\mathbf{d}$, optical flow $\mathbf{o}$, and scene flow $\mathbf{s}$ given pairs of temporally adjacent images as input. 
%

\subsection{Optical Flow Estimation}
\label{sec:optflow}

Following RAFT~\cite{raft}, a shared RGB encoder $g_\theta: \mathbb{R}^{H \times W \times 3} \rightarrow \mathbb{R}^{H/8 \times W/8 \times 256}$ extracts $256$-dimensional features from images $I_t$ and $I_{t+1}$ at $1/8$ the original resolution, and an identical but separate RGB context encoder $g_{\theta'}$ extracts features only from $I_{t+1}$. A correlation layer constructs a $W/8 \times H/8 \times W/8 \times H/8$ 4D correlation volume $C_t^{t+1}$ by taking all pair-wise inner products from feature vectors $g_\theta(I_t)$ and $g_\theta(I_{t+1})$. In practice, we use the same multi-scale pooling strategy as~\cite{raft} to build a correlation pyramid with 4 layers. A GRU-based \cite{gru} update operator then uses the current optical flow estimate (initialized at zero) to look up matching values from $C_t^{t+1}$, and iteratively refines it by calculating a residual value to be added to the current estimate, using the context features $g_{\theta'}(I_t)$ as additional input at each iterative step.

\subsection{Depth Triangulation}
\label{sec:triangulation}

This predicted optical flow is used to generate a depth map via triangulation \cite{deep_mv_depth}. For every pixel $\mathbf{x}_i^t = [x_i^t,y_i^t,1]^T$ in frame $t$, its predicted optical flow corresponds to the displacement $\hat{\mathbf{o}}_i$ between frames $I_t$ and $I_{t+1}$, such that $\mathbf{x}_i^{t+1} = \mathbf{x}_i^t + \hat{\mathbf{o}}_i$.  From this correspondence between pixels, relative rotation $\mathbf{R}_t^{t+1}$ and translation $\mathbf{t}_t^{t+1}$ between frames, and camera intrinsics $\mathbf{K}$, we calculate a \emph{triangulated depth map} $\bar{D}^t = \{\bar{d}_i^t, \ i \in [1;H] \times [1;W] \}$ by solving the following least-squares problem:
\begin{equation}
\bar{d}_i^t = \mathrm{argmin}_{d_i^t} \Big|\Big| \mathbf{x}_i^{t+1} \times \mathbf{K} \Big( \mathbf{R}_t^{t+1} \Big(d_i^t \mathbf{K}^{-1}\mathbf{x}_i^t \Big) + \mathbf{t}_t^{t+1} \Big) \Big|\Big|^2 , 
\label{eq:triangulation}
\end{equation}
where $\mathbf{X}_i = d_i^t \mathbf{K}^{-1}\mathbf{x}_i^t$ is the 3D point (in camera coordinate frame at time $t$) that projects to pixels $\mathbf{x}_i^t$ and $\mathbf{x}_i^{t+1}$ in frames $I_t$ and $I_{t+1}$ respectively, and $\times$ denotes cross-product.
A diagram and example depth map obtained by triangulation is shown in Figure \ref{fig:optflow_depth}.
While these depth maps are quite sharp and consistent when modeling static objects, they completely fail in the presence of dynamic objects due to the static world assumption required for triangulation (warping based on ego-motion $\mathbf{T}_t^{t+1}$). Our proposed DRAFT architecture leverages this initial triangulated depth estimate for subsequent joint estimation of depth and scene flow, as described next. 

\subsection{Depth and Scene Flow Estimation}
\label{sec:depthscnflow}

The depth and scene flow estimation stage uses the same input information as the optical flow stage, i.e. the correlation pyramid $C_{t}^{t+1}$ and the context features $g_{\theta'}(I_t)$, plus the triangulated depth map $\bar{D}_t$ calculated from the predicted optical flow $\mathbf{\hat{o}}$ in Eq.~\eqref{eq:triangulation}. This triangulated depth map is first processed using a depth context encoder $g_{\theta''}(\bar{D}_t)$ and concatenated to the RGB context features. Similarly to the optical flow stage, a GRU recurrent network is used to iteratively refine depth and scene flow predictions. However, since here optical flow is not obtained directly, depth and scene flow predictions must first be projected onto the image plane before looking up matches in the correlation pyramid:
\begin{equation}
\label{eq:projoptflow}
    \mathbf{\hat{o}}_i^{mot} = \mathbf{K}\mathbf{T}_t^{t+1}(\hat{d}_i^t \mathbf{K}^{-1} \mathbf{x}_i^t + \mathbf{\hat{s}}_i^t) - \mathbf{x}_i^t
\end{equation}
where $\hat{d}_i^t$ and $\mathbf{\hat{s}}_i^t$ are respectively depth and scene flow estimates for pixel $i$ in frame $I_t$.
Furthermore, we also use optical flow predictions as initial estimates for correlation pyramid querying. As shown in our ablations (Table \ref{table:kitti_ablation}), this second stage initialization significantly improves depth and scene flow performance. 

\input{figures/masks}

\subsection{Forward and Backward Estimates}
\label{sec:fwdbwd}

To account for occluded regions between consecutive frames, we follow~\cite{jonschkowski2020matters} and use forward-backward consistency of predicted optical flow. Specifically, we use forward flow between frames $I_t \rightarrow I_{t+1}$ and backward flow between $I_{t+1} \rightarrow I_t$ to find inconsistent regions, as denoted in Figure~\ref{fig:masks}. We use these masks to filter out parts of the self-supervised and consistency losses, as described in Section~\ref{sec:trainlosses}, and we ablate their effect in Table~\ref{table:kitti_ablation}. While computing these masks requires two forward passes of the model, this can be done efficiently by reutilizing the transposed features of the correlation pyramid between passes.

\section{Learning with Synthetic Supervision and Real-World Self-Supervision}
\label{sec:trainlosses}

Pre-training on synthetic datasets is a core component of optical and scene flow estimation methods \cite{selfsceneflow,multi-scene-flow,raft,raft3d,jonschkowski2020matters}, due to the difficulty in obtaining real-world ground-truth labels, and recently has also been explored for depth estimation~\cite{guda,gasda,sharingan,virtualworld}. In this work we use a mixed-batch training approach, which has been shown to improve self-supervised domain transfer~\cite{guda}. At each step, we sample real and synthetic batches $\mathcal{B}_R$ and $\mathcal{B}_S$, which are processed independently to generate corresponding real and synthetic losses. Formally, we minimize $\mathcal{L} = \mathcal{L}_{R} + \lambda_S \mathcal{L}_{S}$, with $\mathcal{L}_S$ containing the supervised, self-supervised and consistency losses described in this section, and $\mathcal{L}_R$ containing only self-supervised and consistency losses.  $\lambda_S$ is used to balance the contribution of the two loss components. The total loss is defined as:
\begin{align}
\mathcal{L} 
&= \mathcal{L}_{self} (\mathcal{B}_R) + \mathcal{L}_{const} (\mathcal{B}_R)
\\ 
&+ \lambda_S \Big(  \mathcal{L}_{sup} (\mathcal{B}_S) +  \mathcal{L}_{self} (\mathcal{B}_S) + \mathcal{L}_{const} (\mathcal{B}_S) \Big)
\end{align}

\subsection{Supervised Losses}
\label{sec:suplosses}

The supervised loss is defined as $\mathcal{L}_{sup} = \mathcal{L}_{depth} + \mathcal{L}_{opt} + \mathcal{L}_{scn} + \mathcal{L}_{nrm}$. 
Note that this loss is imposed only on synthetic data, where ground truth is available. 

\noindent\textbf{Depth.} We use the Smooth L1 loss (i.e. the Huber loss) to supervise depth when ground-truth is available.
%
%

\noindent\textbf{Surface Normal Regularization.} Following \cite{guda}, when supervising on synthetic datasets, we leverage dense annotations and apply an additional surface normal regularization term on the predicted depth maps. For any pixel $\mathbf{x} =[u,v] \in D$, its surface normal vector $\mathbf{n} \in \mathbb{R}^3$ can be calculated as:
$\mathbf{n} = ( \mathbf{X}_{u+1,v} - \mathbf{X}_{u,v} ) \times ( \mathbf{X}_{u,v+1} - \mathbf{X}_{u,v} )$, where $\mathbf{X} = d \mathbf{K}^{-1} \mathbf{x}$ is its reconstructed 3D point. As a measure of similarity between ground-truth $\mathbf{n}$ and predicted $\hat{\mathbf{n}}$ surface normal vectors we use the \emph{cosine similarity} metric, defined as:
\begin{equation}
\mathcal{L}_{nrm} = \frac{1}{2V} \sum_{\mathbf{x} \in D} \Big( 1 - \frac{\hat{\mathbf{n}} \cdot \mathbf{n}}{||\hat{\mathbf{n}}|| \hspace{0.2em} ||\mathbf{n}||} \Big)
\end{equation}

\noindent\textbf{Optical and Scene Flow.} We use the L1 loss to supervise optical flow and scene flow when ground-truth is available.


\subsection{Self-Supervised Losses}
\label{sec:supp_selfsuploss}
The self-supervised loss is defined as $\mathcal{L}_{self} = \mathcal{L}_{photo} + \mathcal{L}_{smth}$, i.e., a photometric loss applied to the reprojection error between real and synthesized views, and a smoothness regularization on the predicted depth maps. Below we describe each of these losses in detail:

\noindent\textbf{Photometric Error.} Given a target $I_t$ and reference $I_{t+1}$ images, it is possible to generate a synthesized version of $I_t$ by projecting information from $I_{t+1}$ between viewpoints. This projection can be done using either optical flow predictions or 3D motion predictions (depth and scene flow), such that $\hat{I}_t^{opt} = I_{t+1} \Big\langle \mathbf{x}_i^t + \hat{\mathbf{o}}_i \Big\rangle$ and $\hat{I}_t^{mot} = I_{t+1} \Big\langle \mathbf{K}\mathbf{T}_t^{t+1} (\hat{d}_t \mathbf{K}^{-1} \mathbf{x}_i^t + \hat{\mathbf{s}}_t) \Big\rangle$. 
The $\langle \rangle$ symbol represents a \emph{bilinear sampling operator}, that is locally sub-differentiable and thus can be used as part of an optimization pipeline~\cite{jaderberg2015spatial}.
To measure the reconstruction error we use the standard photometric loss~\cite{photo_loss} consisting of an SSIM component~\cite{wang2004image} and the L1 distance in pixel space. 
The total photometric loss is defined as:
\begin{equation}
\small
\mathcal{L}_{photo} = \mathcal{L}_{photo}(I_t, \hat{I}_t^{opt}) + \mathcal{L}_{photo}(I_t, \hat{I}_t^{mot})
\end{equation}
\noindent\textbf{Smoothness Regularization.} To enforce neighborhood consistency in the predicted depth maps, we use an edge-aware smoothness loss \cite{godard2017unsupervised} on the mean-normalized depth map $\hat{D}_t^* = \hat{D}_t / \bar{D}_t$. This loss enforces gradient smoothness in areas with low texture, while allowing for depth discontinuities in high-textured areas (i.e., object boundaries).


\subsection{Consistency Losses}
\label{sec:supp_consistencylosses}

The consistency loss is defined as $\mathcal{L}_{const} = \mathcal{L}_{mot}^{opt} + \mathcal{L}_{rev}^{opt} + \mathcal{L}_{rev}^{mot} + \mathcal{L}_{reproj}^{depth}$, respectively constraining learned optical flow to be similar to projected depth and scene flow; backward/forward optical flow consistency; and backward/forward reprojected depth consistency. 

\noindent\textbf{Optical flow.} As discussed in Section \ref{sec:depthscnflow}, in addition to directly estimating optical flow, we can also calculate it indirectly by projecting depth and scene flow estimates onto the image plane (Eq. \ref{eq:projoptflow}). Constraining these two predictions to be the same provides an additional source of regularization for the ill-posed problem of depth and scene flow estimation in a monocular setting.
\begin{equation}
\small
\label{eq:consistency1}
    \mathcal{L}_{opt}^{mot} = \frac{1}{HW} \sum \Big| \mathbf{\hat{o}}_i - ( \mathbf{K}\mathbf{T}_t^{t+1}(\hat{d}_t \mathbf{K}^{-1} \mathbf{x}_i^t + \hat{\mathbf{s}}_t) - \mathbf{x}_i^t) \Big|
\end{equation}

\noindent\textbf{Reverse Optical Flow.}
Another source of optical flow regularization comes from temporal consistency, i.e. the predicted forward optical flow should be the same as the back-projected predicted backward optical flow (and vice-versa). This back-projected optical flow can be efficiently computed using the same warping coordinates calculated for the reprojection error, however now instead of warping RGB values we are warping optical flow predictions. Note that, since the reverse optical flow is calculated, the \emph{sum} of predictions is to be minimized, and not the difference. The same process can also be applied to the predicted optical flow calculated from motion (i.e., depth and scene flow), for an additional source of regularization.
\begin{align}
\small
\label{eq:consistency2a}
\mathcal{L}_{rev}^{opt} &= \frac{1}{HW} \sum \Big|
\hat{O}_t + \hat{O}_{t+1}^{opt} \Big\langle \mathbf{x}_i^t +  \hat{\mathbf{o}}_i \Big\rangle \Big|
\\
\label{eq:consistency2b}
\mathcal{L}_{rev}^{mot} &= \frac{1}{HW} \sum \Big| \hat{O}_t^{mot} + \hat{O}_{t+1}^{mot} \Big\langle \mathbf{K}\mathbf{T}_t^{t+1} (\hat{d}_t \mathbf{K}^{-1} \mathbf{x}_i^t + \hat{\mathbf{s}}_t) \Big\rangle \Big|
\end{align}

\noindent\textbf{Reprojected Depth.}
Similarly, we can also temporally regularize depth estimations by back-projecting predictions from one timestep onto another and enforcing similarity. Note that, because depth changes with viewpoint, grid-sampling for depth reconstruction needs to be performed instead on $\tilde{D}_{t} = \Big| \mathbf{T}_t^{t+1}(\hat{D}_{t+1} \mathbf{K}^{-1} \mathbf{x}_i^t + \hat{S}_t) \Big|_z$, i.e. the back-projected range values from $\hat{D}_{t+1}$.
\begin{equation}
\small
\label{eq:consistency3}
\mathcal{L}_{reproj}^{depth} = \frac{1}{HW} \sum \Big| \hat{D}_t - \tilde{D}_{t} \Big\langle \mathbf{K}\mathbf{T}_t^{t+1} (\hat{d}_t \mathbf{K}^{-1} \mathbf{x}_i^t + \hat{\mathbf{s}}_t) \Big\rangle \Big|
\end{equation}
\vspace{-8mm}

%% file: figures/optflow_depth.tex
\begin{figure*}[t!]
    \centering
\begin{tabular}{ccc}
\multirow{2}{*}{
\raisebox{0mm}[28mm]{
\subfloat[Two-frame triangulation]{
\label{fig:triangulation}
\includegraphics[width=0.3\textwidth]{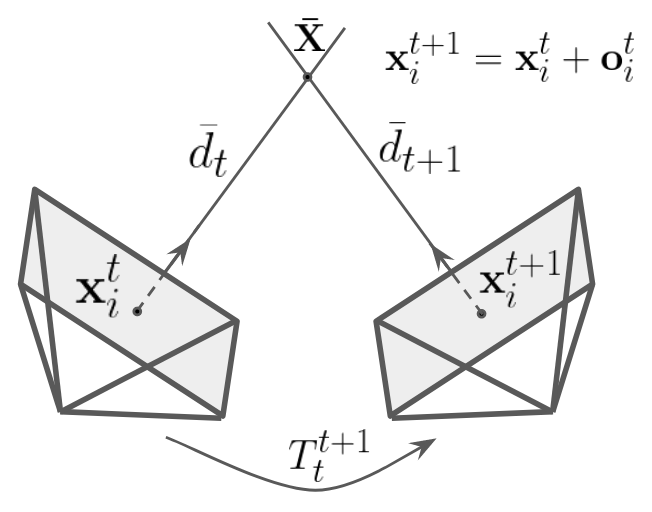}
}}}
&
\vspace{2mm}
\hspace{-4mm}
    \subfloat[RGB image]{
    \includegraphics[width=0.3\textwidth]{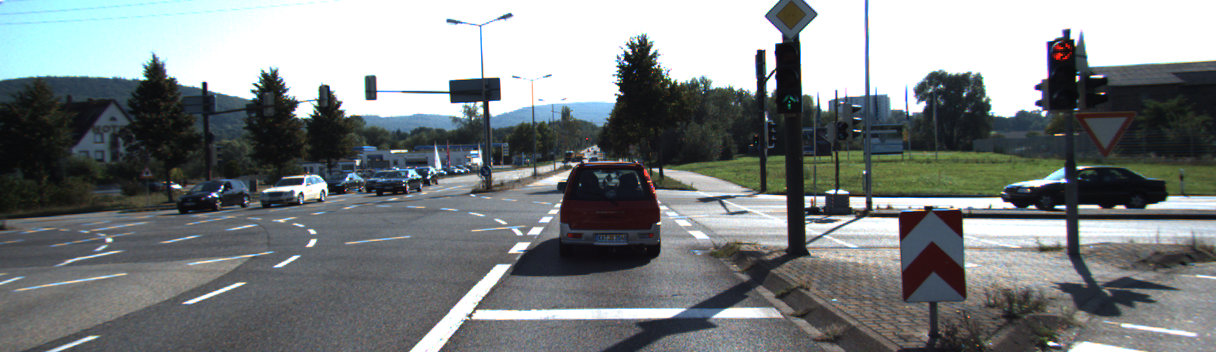}
    }
&
\hspace{-4mm}
    \subfloat[Predicted optical flow]{
    \includegraphics[width=0.3\textwidth]{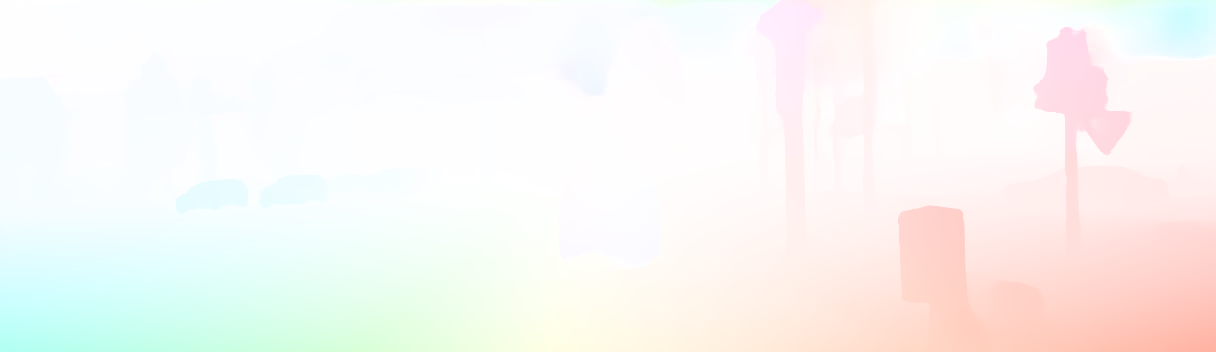}
    }
\\ 
&
\hspace{-4mm}
    \subfloat[Triangulated depth map]{
    \includegraphics[width=0.3\textwidth]{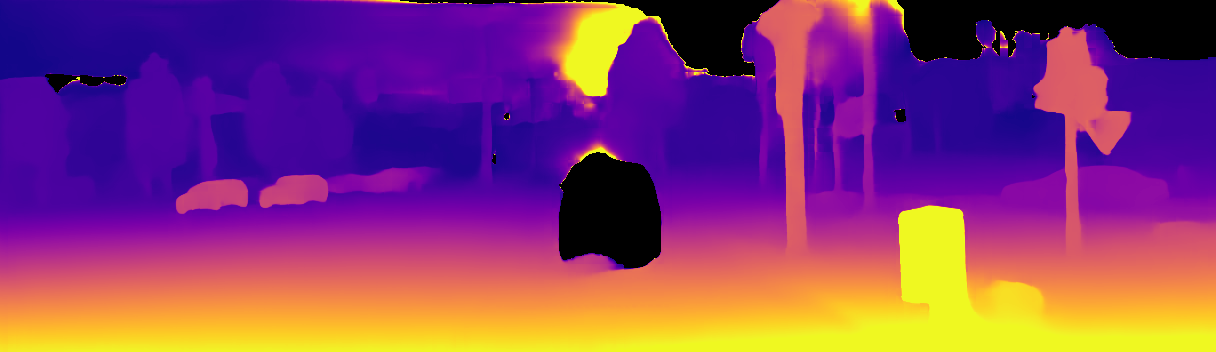}
    }
&    
\hspace{-4mm}
    \subfloat[Predicted depth map]{
    \includegraphics[width=0.3\textwidth]{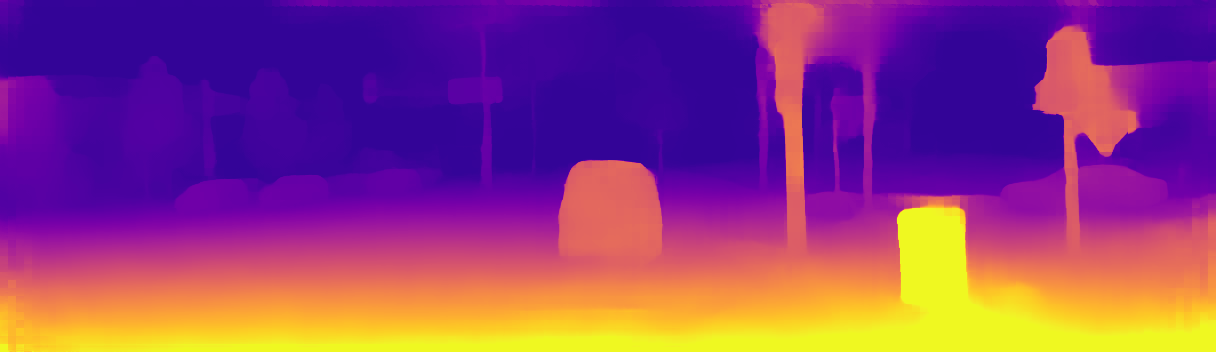}
    }

    \end{tabular}
    \caption{\textbf{Example of triangulated depth map generated from optical flow estimates}, on the KITTI dataset. The presence of moving objects violates the static world assumption required for triangulation, leading to poor performance on moving objects. Our DRAFT architecture uses this initial estimate as additional input to a depth and scene flow network for refinement.}
    \label{fig:optflow_depth}
\vspace{-5mm}
\end{figure*}

%% file: figures/masks.tex

\begin{figure}[t!]
    \centering
    \subfloat{
    \includegraphics[width=0.22\textwidth]{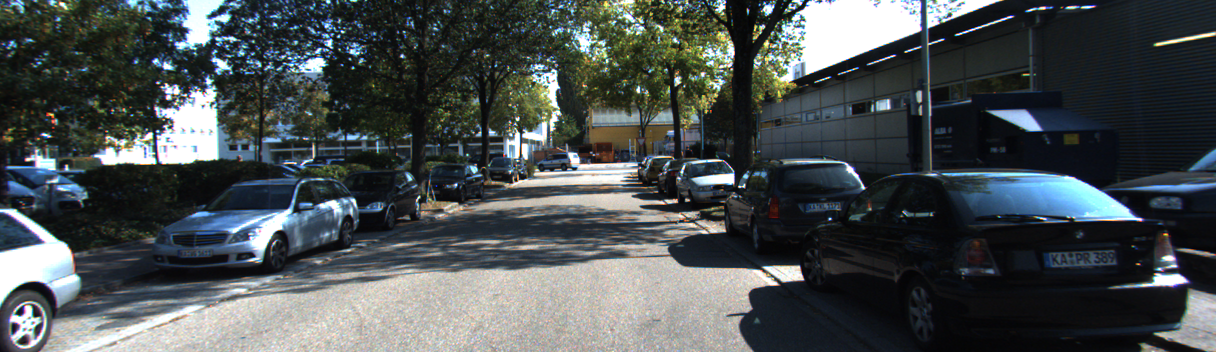}
    }
    \hspace{-2mm}
    \subfloat{
    \includegraphics[width=0.22\textwidth]{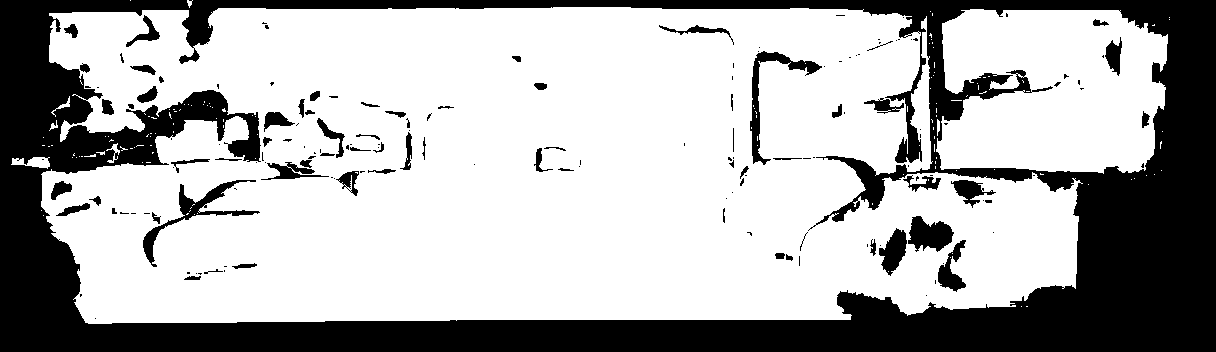}
    } \\ \vspace{-3mm}
    \subfloat{
    \includegraphics[width=0.22\textwidth]{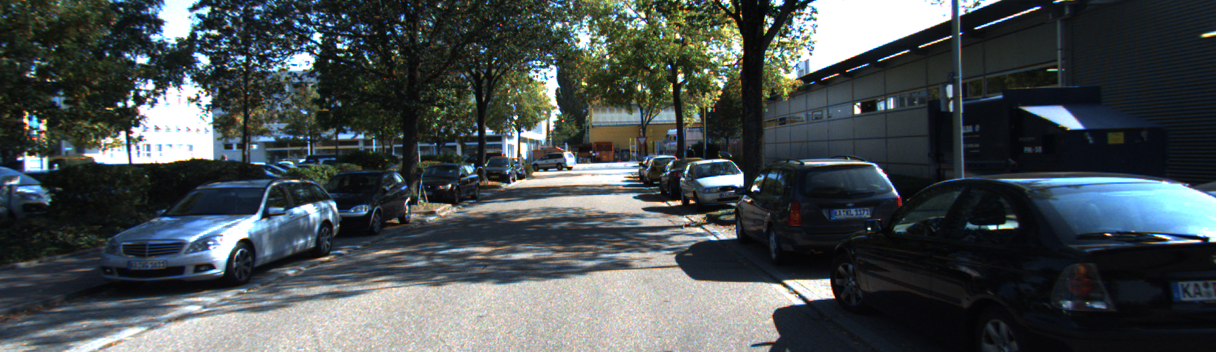}
    }
    \hspace{-2mm}
    \subfloat{
    \includegraphics[width=0.22\textwidth]{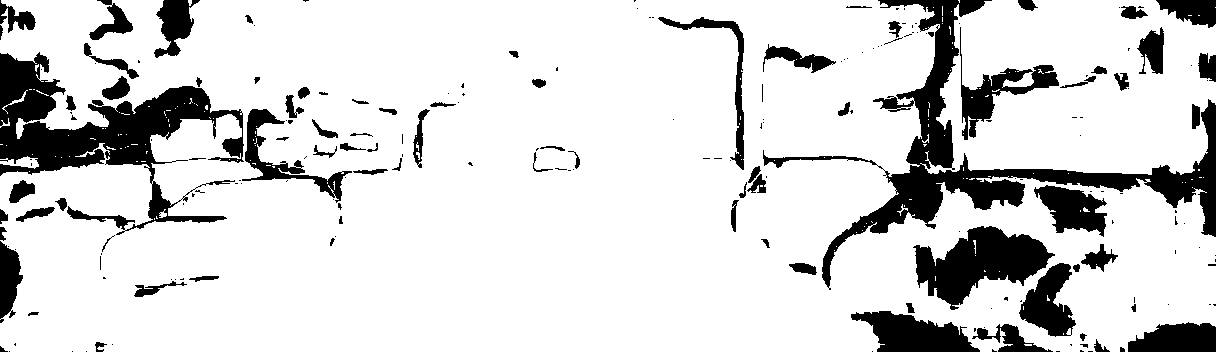}
    }
\caption{\textbf{Occlusion masks} obtained from backwards and forwards predicted optical flow. These masks are used to filter information from self-supervised and consistency losses.}
    \label{fig:masks}
\end{figure}

%% file: sections/04experiments.tex
\vspace{2mm}
\section{Experimental Results}

\input{tables/kitti_depth}

\input{tables/kitti_optscnflow}

\subsection{Datasets}

\textbf{KITTI~\cite{geiger2013vision}}
is the standard benchmark for optical flow, depth and scene flow evaluation. Following standard protocol, we use the \textit{Eigen} split with the motion filtering of~\cite{zhou2016learning} as the source or real-world self-supervision, resulting in $39810$ training images. For depth evaluation, we use the \textit{Eigen} test split~\cite{eigen2014depth}, containing $697$ images with corresponding ground-truth depth maps. For optical flow and scene flow evaluation, we use the official $200$ annotated training frames~\cite{Menze2015ISA}.

\textbf{VKITTI2~\cite{cabon2020vkitti2}}
is a more photo-realistic version of the original VKITTI dataset~\cite{gaidon2016virtual}, containing reconstructions of five sequences of the KITTI odometry benchmark~\cite{Geiger2012CVPR} with corresponding dense depth maps, camera motion, optical flow and scene flow annotations. Importantly, this dataset contains similar camera extrinsics and intrinsics as KITTI, which enables the direct scale transfer between datasets. We used all forward-facing sequences from camera $0$, including the \emph{clone}, \emph{fog}, \emph{morning}, \emph{overcast}, \emph{rain} and \emph{sunset} variations, for a total of $12936$ samples.

\input{tables/kitti_ablation}

\textbf{Parallel Domain~\cite{parallel_domain}}
is a procedurally-generated dataset and contains fully annotated photo-realistic renderings of urban driving scenes, including multiple cameras and LiDAR sensors.  It contains $40000$ samples with corresponding dense depth maps, camera motion, optical flow and scene flow annotations. We use this dataset as an additional source of synthetic supervision in our experiments, showing significant improvements when combined with  VKITTI2.

\subsection{Experimental Protocol}
\label{sec:protocol}

Our models were implemented using PyTorch~\cite{paszke2017automatic} and trained across 8 GPUs. At each training step, samples from a synthetic and real-world dataset are randomly selected and processed using the same architecture, using $12$ recurrent update steps. Other network parameters were taken from~\cite{raft}: $4$ correlation levels with lookup radius of $4$, $256$ channels for RGB features, $128$ for context RGB and depth features, and $128$ for update blocks. Each sample is processed twice, with inverted inputs to generate backward and forward predictions (Sec. \ref{sec:fwdbwd}). The losses from each sample (Sec. \ref{sec:trainlosses}) are used to perform a single gradient back-propagation step, using the Adam optimizer~\cite{kingma2014adam}, with $\beta_1=0.9$ and $\beta_2=0.999$. We use the following loss weight coefficients, determined via one-dimensional grid search: $\lambda_{depth} = 5.0$, 
$\lambda_{opt} = 0.1$, 
$\lambda_{scn} = 2.0$, 
$\lambda_{nrm} = 0.05$, 
$\lambda_{reproj} = 1.0$, 
$\lambda_{smth} = 0.0001$, 
$\lambda_{mot}^{opt} = 0.2$, 
$\lambda_{rev}^{opt} = \lambda_{rev}^{mot} = 0.1$, and
$\lambda_{reproj}^{depth} = 0.005$. 

\textcolor{black}
{Starting from random weights, we train for 30 epochs on a combination of Parallel Domain and KITTI, with an initial learning rate of $10^{-4}$ that is halved at epochs $20$ and $25$. We then train for another $30$ epochs on a combination of VKITTI2 and KITTI, with initial learning rate $5 \cdot 10^{-5}$, again halved at epochs $20$ and $25$. As shown in our ablation studies (Table \ref{table:kitti_ablation}), training with both synthetic datasets leads to more robust geometric priors that can be directly transferred between domains. This is in accordance with other published works \cite{raft,raft3d,monosf,multi-scene-flow}. Other synthetic datasets were not considered because they do not contain all the geometric tasks we estimate (depth, optical flow and scene flow).  
}

\input{figures/qualitative}

\subsection{Comparison with Other Methods}

We evaluate our proposed architecture on the KITTI dataset for all three considered tasks: \emph{optical flow}, \emph{depth} and \emph{scene flow}, comparing with related self-supervised methods. To the best of our knowledge, DRAFT is the first architecture to simultaneously estimate all three tasks, so we compare with methods that perform only a subset of them. Conversely, our reported results are \emph{obtained with a single model}, trained to perform all three tasks simultaneously.

\input{figures/comparison}

In Table \ref{table:kitti_depth} we report depth estimation results. 
We note that methods that jointly estimate depth and scene flow~\cite{epc,epc++,selfsceneflow} --- even on a stereo setting --- achieve sub-par results relative to methods that perform only depth estimation. 
We attribute this behavior to the fact that depth and scene flow are complementary in the self-supervised setting, and introducing scene flow actually destabilizes depth performance. Our architecture, on the other hand, is able to leverage scene flow predictions in such a way that depth predictions both (1) do not degrade in static scenes; and (2) improve in the presence of dynamic objects (see Fig. \ref{fig:kitti_comparison}). 
Secondly, DRAFT also outperforms other methods that use synthetic data as an additional source of training data for self-supervised monocular depth estimation~\cite{gasda,sharingan,virtualworld}. Importantly, this is achieved without explicit domain adaptation, solely by enforcing a combination of supervised losses in the synthetic domain and self-supervised losses in the real-world domain in a mixed-batch setting. The only method competitive with DRAFT is Manydepth~\cite{manydepth}, a multi-frame architecture for depth estimation that uses cost volumes in combination with a single-frame teacher network. However, the lower Sq.Rel. and RMSE values achieved by DRAFT indicate better reasoning over outliers, such as dynamic objects.

In Table \ref{tab:optscnflow} we report optical flow and scene flow results, obtained using the same model evaluated in Table \ref{table:kitti_depth} for depth estimation. In Table \ref{table:optflow} we show that the self-supervision introduced by DRAFT substantially improves optical flow performance relative to the direct transfer method commonly reported by supervised methods, especially relative to the RAFT~\cite{raft} baseline. Not many methods report self-supervised optical flow results~\cite{jonschkowski2020matters,zou2018dfnet}, and consistently outperform these methods, while also producing state of the art depth and scene flow estimates.  Similarly, in Table \ref{table:scnflow} we show that DRAFT outperforms other self-supervised scene flow methods, even though it operates in the much more challenging monocular setting, that does not use stereo cameras at training time as a way to generate scale-aware models and decouple depth from scene flow estimation. 



\subsection{Ablation Study}

We also performed a detailed ablation of each component, with results reported in Table \ref{table:kitti_ablation}. We ablate the effects of single task learning (\emph{depth only} and \emph{optical flow only}), showing that performance actually degrades in this setting, especially in the case of depth, that does not benefit from optical flow initialization. Similarly, removing scene flow leads to considerably worse root mean squared error (RMSE) depth estimation values, due to outliers generated by the presence of dynamic objects. The removal of task consistency losses (Eq. \ref{eq:consistency1}) has a significant impact in performance --- especially in depth and scene flow --- as well as the removal of temporal consistency losses (Equations \ref{eq:consistency2a}, \ref{eq:consistency2b} and \ref{eq:consistency3}). Removing self-supervised losses on synthetic data has a small but consistent overall impact in performance for all considered tasks. We also ablated the use of occlusion masks to remove spurious information from the self-supervised and consistency losses, showing that they indeed improve performance across the board. Using triangulated depth as additional input to the depth and scene flow networks has a non-trivial impact in these tasks, as well as optical flow initialization. \textcolor{black}{We also experimented training with a single synthetic dataset (VKITTI2 or Parallel Domain), for a total of $30$ epochs as described in Section \ref{sec:protocol}. Although we achieve competitive results with any single dataset, it is clear that using both of them leads to optimal performance, as an additional source of geometric priors to be transferred to the real-world through self-supervision}. Finally, we also experimented training  with only forward or backward information, with marginal degradation relative to the proposed forward-backward methodology.

%% file: tables/kitti_depth.tex
\captionsetup[table]{skip=6pt}

\begin{table*}[t!]
\small
\renewcommand{\arraystretch}{0.9}
\centering
{
\small
\setlength{\tabcolsep}{0.3em}
\begin{tabular}{l|c|c|c|c|cccc|ccc}
\toprule
\multirow{2}[2]{*}{\textbf{Method}} & 
\multirow{2}[2]{*}{\rotatebox{90}{\scriptsize{Superv.}}} &
\multirow{2}[2]{*}{\rotatebox{90}{\scriptsize{Synth.}}} & 
\multirow{2}[2]{*}{\rotatebox{90}{\scriptsize{Multi.}}} & 
\multirow{2}[2]{*}{\rotatebox{90}{\scriptsize{S.Flow}}} & 
\multicolumn{4}{c|}{\textit{Lower is better}} &
\multicolumn{3}{c}{\textit{Higher is better}} 
\\
\cmidrule(lr){6-9} \cmidrule(lr){10-12}
& & & & &
AbsRel &
SqRel &
RMSE &
RMSE$_{log}$ &
$\delta < {1.25}$ &
$\delta < {1.25}^2$ &
$\delta < {1.25}^3$
\\
\midrule
EPC \cite{epc}
& S & & & \checkmark
& 0.127 & 1.239 & 6.247 & 0.214 & 0.847 & 0.926 & 0.969 
\\
EPC++ \cite{epc++}
& S & & & \checkmark
& 0.127 & 0.936 & 5.008 & 0.209 & 0.841 & 0.946 & 0.979 
\\
Self-Mono-SF \cite{selfsceneflow}
& S & & & \checkmark
& 0.125 & 0.978 & 4.877 & 0.208 & 0.851 & 0.950 & 0.978 
\\
GASDA \cite{gasda}
& M & \checkmark & &
& 0.120 & 1.022 & 5.162 & 0.215 & 0.848 & 0.944 & 0.974
\\
SharinGAN \cite{sharingan}
& M & \checkmark & &
& 0.116 & 0.939 & 5.068 & 0.203 & 0.850 & 0.948 & 0.978
\\
Monodepth2$^\dagger$ \cite{monodepth2}
& M & & &
& 0.115 & 0.882 & 4.701 & 0.190 & 0.879 & 0.961 & 0.982
\\
\textcolor{black}{TrianFlow$^\dagger$ \cite{zhao2020towards}}
& M & \checkmark & &
& 0.113 & 0.704 & 4.581 & 0.184 & 0.871 & 0.961 & \textbf{0.984}
\\
PackNet-SfM$^\dagger$ \cite{packnet}
& M &  & &
& 0.107 & 0.802 & 4.538 & 0.186 & 0.889 & 0.962 & 0.981
\\
\textcolor{black}{DeepVO-OA$^\dagger$ \cite{deepvo-oa}}
& M & & &
& 0.106 & \underline{0.701} & \underline{4.129} & 0.210 & 0.889 & \underline{0.967} & \textbf{0.984}
\\
MonoDEVSNet$^\dagger$ \cite{virtualworld}
& M & \checkmark & &
& 0.104 & 0.721 & 4.396 & 0.185 & 0.880 & 0.962 & 0.983
\\
ManyDepth$^\dagger$ \cite{manydepth}
& M & & \checkmark &
& \textbf{0.093} & 0.703 & 4.215 & \underline{0.173} & \textbf{0.907} & 0.966 & \underline{0.983}
\\
\midrule
\textbf{DRAFT} 
& M & \checkmark & \checkmark & \checkmark
& \underline{0.097} & \textbf{0.647} & \textbf{3.991} & \textbf{0.169} & \underline{0.899} & \textbf{0.968} & \textbf{0.984} 
\\
\bottomrule
\end{tabular}
}
\caption{
\textbf{Depth results} on the \emph{Eigen} split \cite{eigen2014depth}, for distances up to 80m with the \emph{Garg} crop. \emph{Superv.} indicates the source of depth self-supervision (S for stereo and M for monocular); \emph{Synth.} the use of synthetic data; \emph{Multi} the use of multiple input frames; and \emph{S.Flow} whether scene flow is also jointly estimated. \textcolor{black}{Methods marked with $^\dagger$ also estimate pose.}
}
\label{table:kitti_depth}
\end{table*}

%% file: tables/kitti_optscnflow.tex
\begin{table*}[t!]
\centering
\subfloat[\textbf{Optical flow results.} DRAFT substantially improves over the baseline of \cite{raft}, and also outperforms other self-supervised (SS) methods.]{%
\label{table:optflow}
\input{tables/kitti_optflow}
}
\qquad
\subfloat[\textbf{Scene flow results.} The \emph{D} column indicates the source of depth estimates: stereo (S) or monocular (M) monocular self-supervision. Even though DRAFT is purely monocular, it still compares favourably to stereo baselines.]{%
\label{table:scnflow}
\input{tables/kitti_scnflow}}
\caption{\textbf{KITTI optical and scene flow results} on the official training images, relative to other self-supervised methods. }
\label{tab:optscnflow}
\end{table*}

%% file: tables/kitti_optflow.tex
\setlength{\tabcolsep}{5pt}
\small
\begin{tabular}{l|c|cc}
\toprule
\renewcommand{\arraystretch}{0.9}
\textbf{Method} & SS &
EPE$\downarrow$ &
F1-all$\downarrow$
\\
\midrule
HD3 \cite{hd3} & &
13.7 & 24.0 
\\
PWC-Net \cite{pwcnet} & &
10.35 & 33.7
\\
FlowNet2 \cite{flownet2} & &
10.08 & 30.0
\\
VCN \cite{vcn} & &
8.36 & 25.1 
\\
DF-Net \cite{zou2018dfnet} & \checkmark &
8.98 & 26.0 
\\
RAFT \cite{raft} & &
5.04 & 17.4 
\\
\textcolor{black}{TrianFlow} \cite{zhao2020towards} & &
3.60 & 18.05 
\\
UFlow \cite{jonschkowski2020matters} & \checkmark &
2.71 & ---
\\
\midrule
\textbf{DRAFT} & \checkmark & 
\textbf{2.54} & \textbf{14.8}
\\
\bottomrule
\end{tabular}

%% file: tables/kitti_scnflow.tex
\setlength{\tabcolsep}{2pt}
\small
\begin{tabular}{l|c|cccc}
\toprule
\textbf{Method} & D &
D1-all$\downarrow$ &
D2-all$\downarrow$ &
F1-all$\downarrow$ &
SF1-all$\downarrow$ 
\\
\midrule
DF-Net \cite{zou2018dfnet} & S
& 46.50 & 61.54 & 27.47 & 73.30
\\
GeoNet \cite{yin2018geonet} & S
& 49.54 & 58.17 & 37.83 & 71.32
\\
EPC \cite{epc} & S
& 26.81 & 60.97 & 25.74 & ---
\\
EPC++ \cite{epc++} & S
& \textbf{23.84} & 60.32 & 19.64 & ---
\\
Self-Mono-SF \cite{selfsceneflow} & S
& 31.25 & 34.86 & 23.49 & 47.05
\\
Multi-Mono-SF \cite{multi-scene-flow} & S
& 27.33 & \underline{30.44} & \underline{18.92} & \underline{39.82}
\\
\midrule
\textbf{DRAFT} & M 
& \underline{26.41} & \textbf{28.89} & \textbf{18.71} & \textbf{37.58} 
\\
\bottomrule
\end{tabular}


%% file: tables/kitti_ablation.tex
\captionsetup[table]{skip=6pt}

\begin{table*}[t!]
\renewcommand{\arraystretch}{0.9}
\centering
{
\small
\setlength{\tabcolsep}{0.6em}
\begin{tabular}{l|ccc|cc|cc}
\toprule
\multirow{2}{*}{\textbf{Method}} &
\multicolumn{3}{c|}{\textit{Depth}} &
\multicolumn{2}{c|}{\textit{Opt. Flow}} &
\multicolumn{2}{c}{\textit{Scene Flow}}
\\
\cmidrule(lr){2-4} \cmidrule(lr){5-6} \cmidrule(lr){7-8}
&
AbsRel &
RMSE &
$\delta < {1.25}$ &
EPE &
F1-all &
F1-all &
SF1-all
\\
\midrule
Direct transfer
& 0.152 & 5.744 & 0.793 & 5.41 & 21.9 & 32.4 & 83.9  
\\
\midrule
Depth only
& 0.108 & 4.449 & 0.865 & --- & --- & --- & ---  
\\
Optical flow only
& --- & --- & --- & 2.88 & 15.1 & --- & --- 
\\
W/o scene flow 
& 0.098 & 4.208 & 0.892 & 2.75 & 15.3 & --- & --- 
\\
\midrule
W/o task consistency 
& 0.106 & 4.347 & 0.885 & 2.72 & 15.2 & 20.45 & 40.98  
\\
W/o temporal consistency 
& 0.103 & 4.216 & 0.888 & 2.83 & 15.7 & 20.37 & 39.77  
\\
W/o synth. self-supervision
& 0.099 & 4.057 & 0.896 & 2.68 & 15.1 & 19.32 & 38.21  
\\
\midrule
W/o occlusion masks
& 0.104 & 4.119 & 0.891 & 2.78 & 15.4 & 21.49 & 41.28  
\\
W/o depth triangulation 
& 0.107 & 4.303 & 0.878 & 2.64 & 15.0 & 20.88 & 41.12  
\\
W/o optflow initialization 
& 0.102 & 4.142 & 0.882 & 2.66 & 14.9 & 20.01 & 39.66  
\\
\midrule
VKITTI2 only 
& 0.105 & 4.461 & 0.874 & 3.13 & 16.7 & 22.31 & 43.55  
\\
\textcolor{black}{Parallel Domain only}
& 0.103 & 4.404 & 0.851 & 3.38 & 16.9 & 21.14 & 42.36  
\\
\midrule
Forward only 
& 0.099 & 4.198 & 0.891 & 2.80 & 15.1 & 20.26 & 38.41  
\\
Backward only 
& 0.101 & 4.076 & 0.885 & 2.98 & 15.7 & 20.97 & 39.94  
\\
\midrule
\textbf{DRAFT} 
& \textbf{0.097} & \textbf{3.991} & \textbf{0.899} & \textbf{2.54} & \textbf{14.8} & \textbf{18.71} & \textbf{37.58}  
\\


\bottomrule
\end{tabular}
}
\caption{
\textbf{DRAFT ablation analysis} on the KITTI dataset. We ablate variations such as single task learning, the removal of different losses, not using occlusion masks, depth triangulation, or optical flow initialization, the removal of \emph{Parallel Domain} as an additional synthetic dataset, and using only forward or backward information during training.
}
\label{table:kitti_ablation}
\vspace{-5mm}
\end{table*}

%% file: figures/qualitative.tex
\begin{figure*}[t!]
\vspace{-3mm}
    \centering
    \subfloat{
    \includegraphics[width=0.24\textwidth]{images/results/val-KITTI_raw-eigen_val_files-velodyne-760-rgb.png}}
    \subfloat{
    \includegraphics[width=0.24\textwidth]{images/results/val-KITTI_raw-eigen_val_files-velodyne-760-pred_fwd_optical_flow.png}}
    \subfloat{
    \includegraphics[width=0.24\textwidth]{images/results/val-KITTI_raw-eigen_val_files-velodyne-760-inv_depths.png}}
    \subfloat{
    \includegraphics[width=0.24\textwidth]{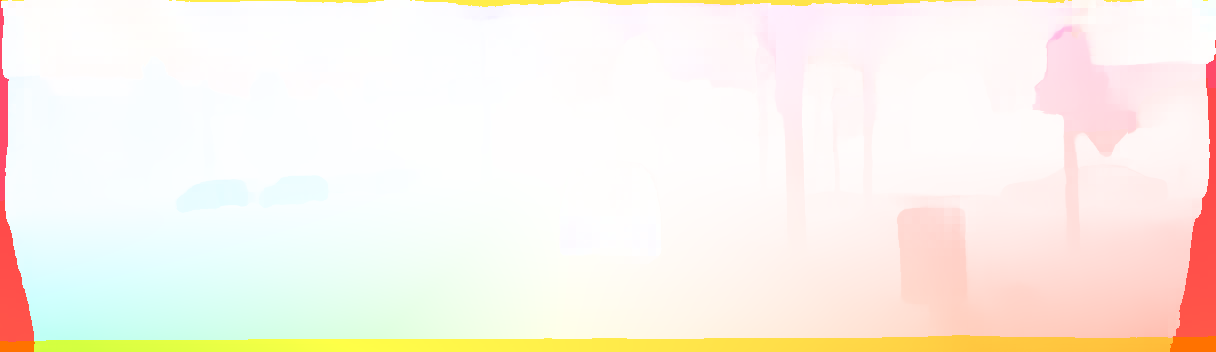}}
    \\
    \vspace{-3.5mm}
    \subfloat{
    \includegraphics[width=0.24\textwidth]{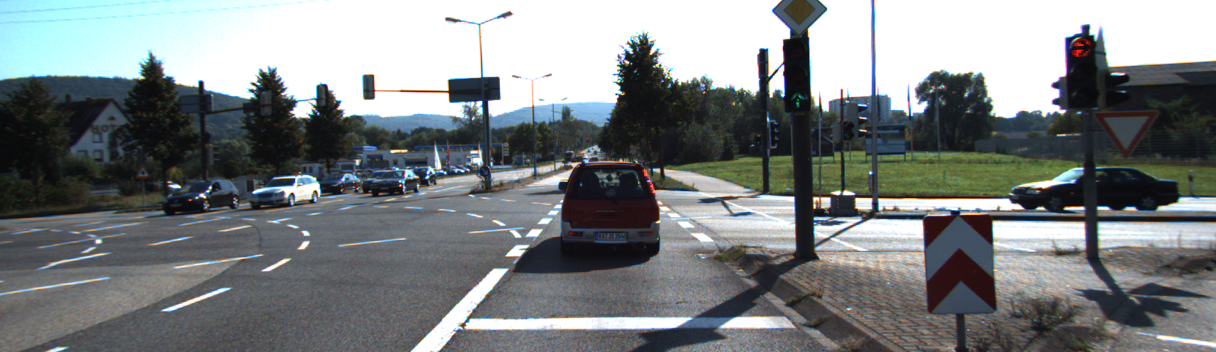}}
    \subfloat{
    \includegraphics[width=0.24\textwidth]{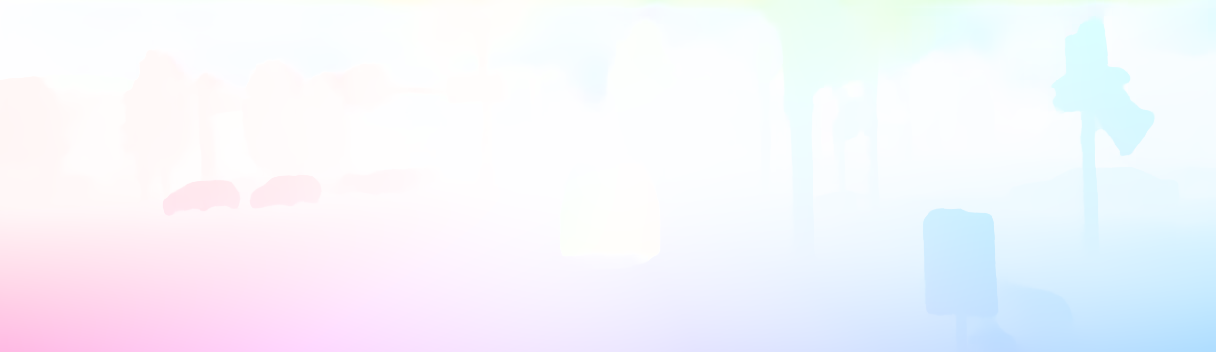}}
    \subfloat{
    \includegraphics[width=0.24\textwidth]{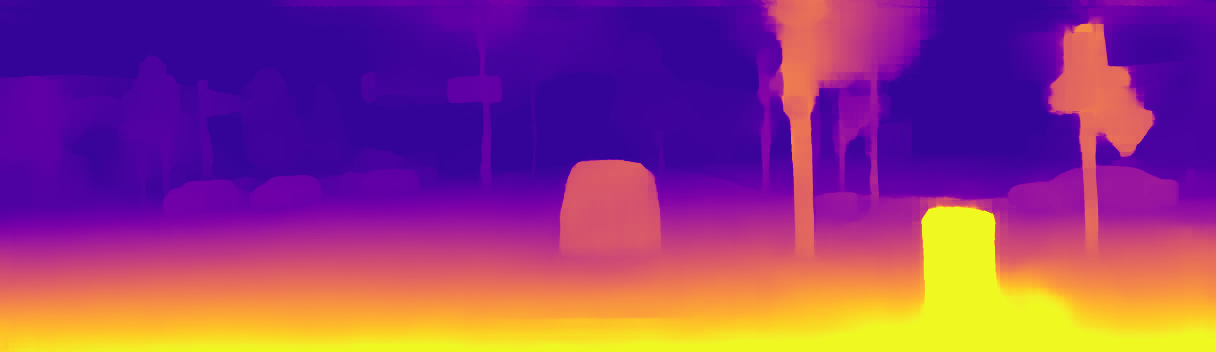}}
    \subfloat{
    \includegraphics[width=0.24\textwidth]{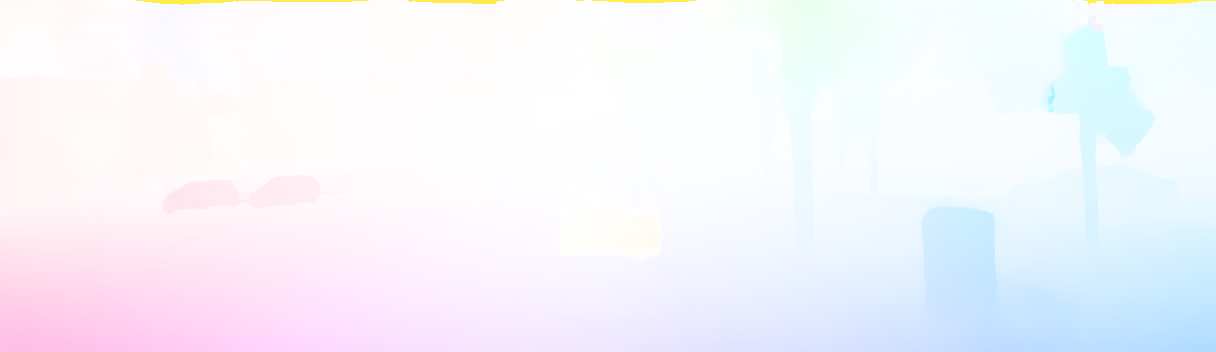}}
    \\    
    \vspace{-2.0mm}
    \subfloat{
    \includegraphics[width=0.24\textwidth]{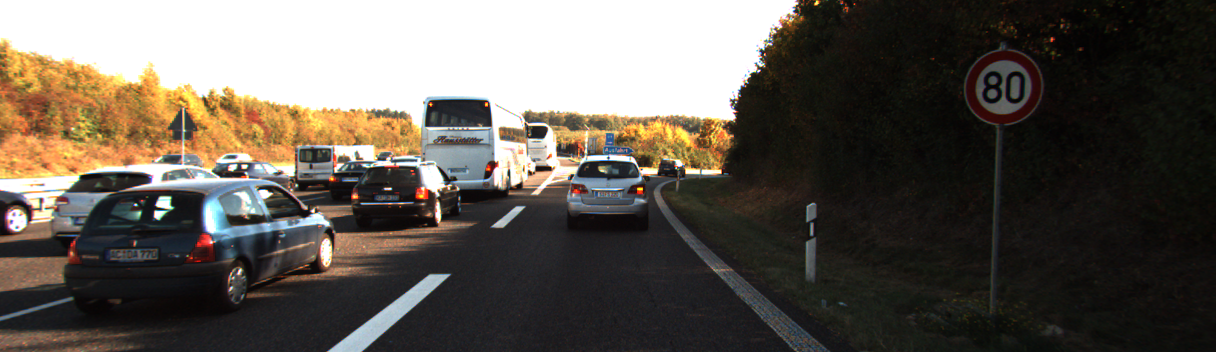}}
    \subfloat{
    \includegraphics[width=0.24\textwidth]{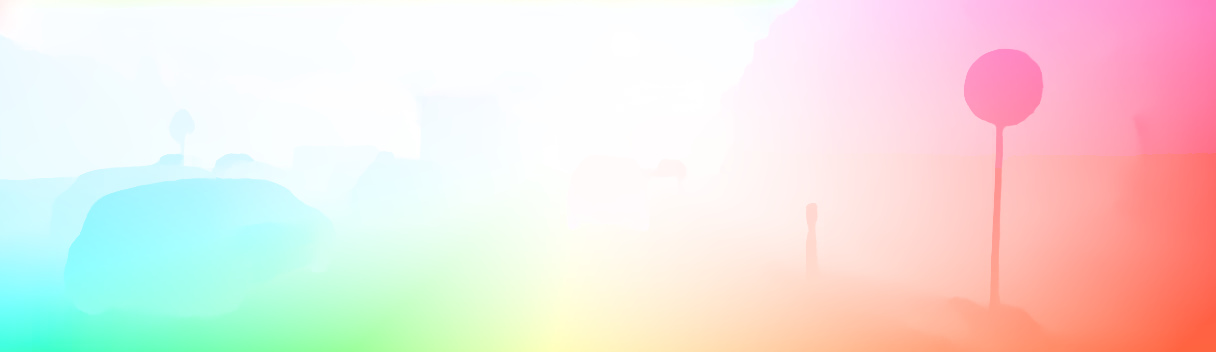}}
    \subfloat{
    \includegraphics[width=0.24\textwidth]{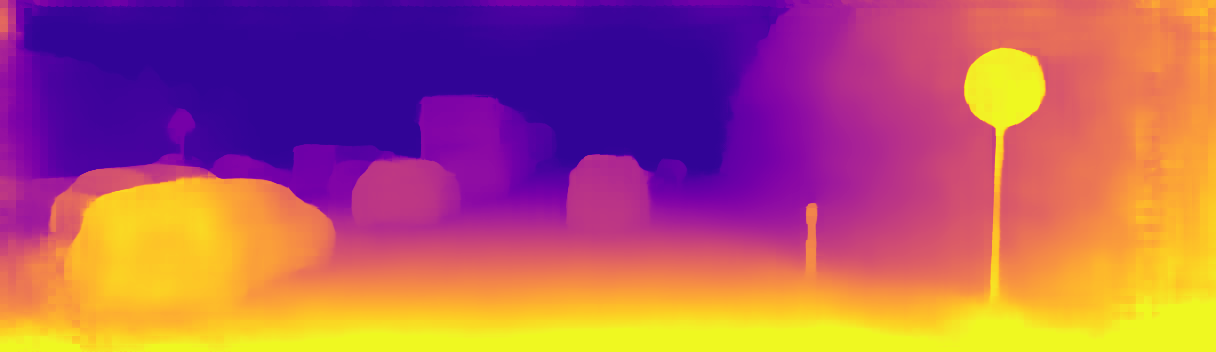}}
    \subfloat{
    \includegraphics[width=0.24\textwidth]{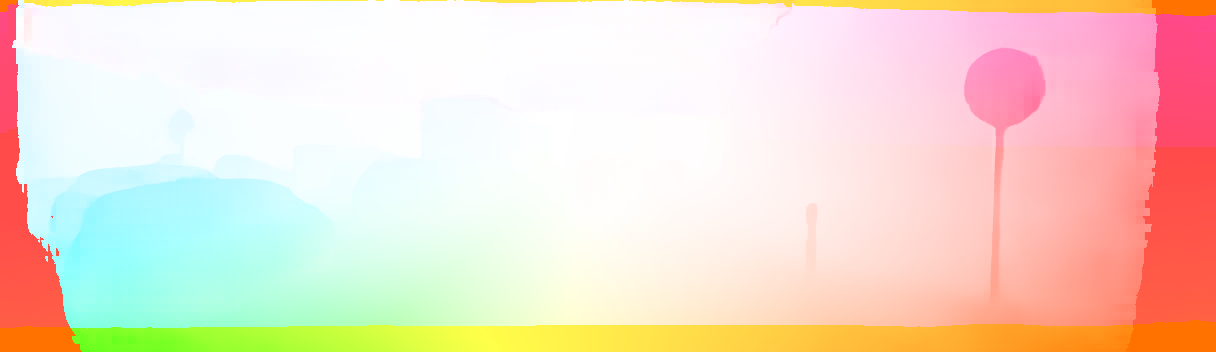}}
    \\
    \setcounter{subfigure}{0}
    \vspace{-3.5mm}
    \subfloat[Input images]{
    \includegraphics[width=0.24\textwidth]{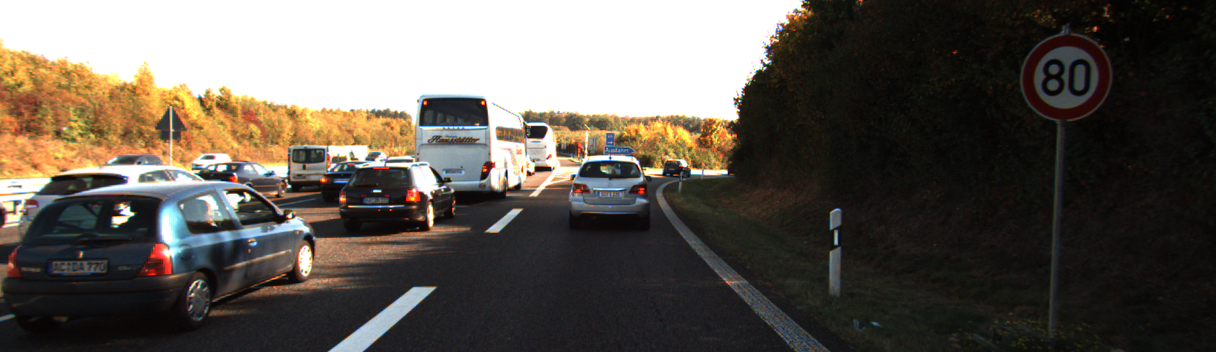}}
    \subfloat[Predicted optical flow]{
    \includegraphics[width=0.24\textwidth]{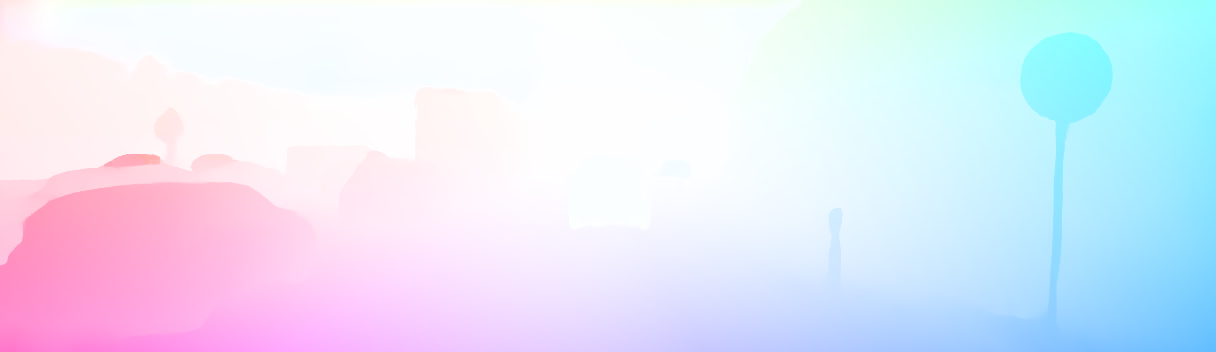}}
    \subfloat[Predicted depth]{
    \includegraphics[width=0.24\textwidth]{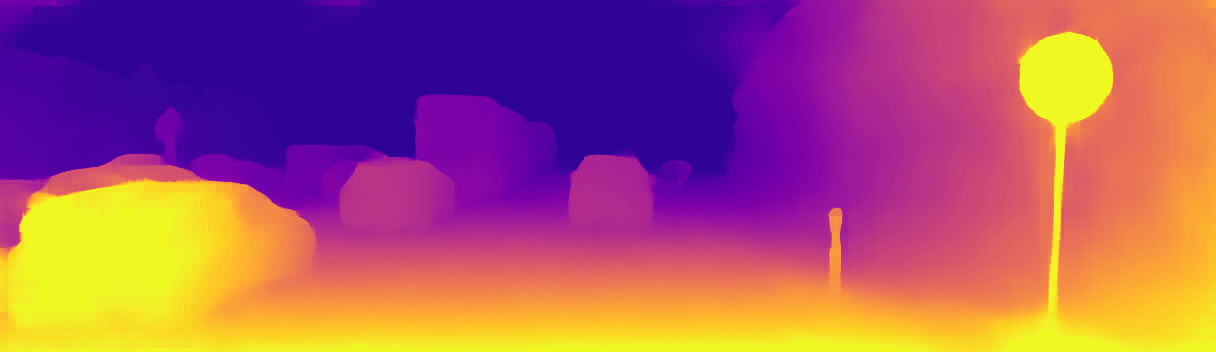}}
    \subfloat[Predicted projected optical flow]{
    \includegraphics[width=0.24\textwidth]{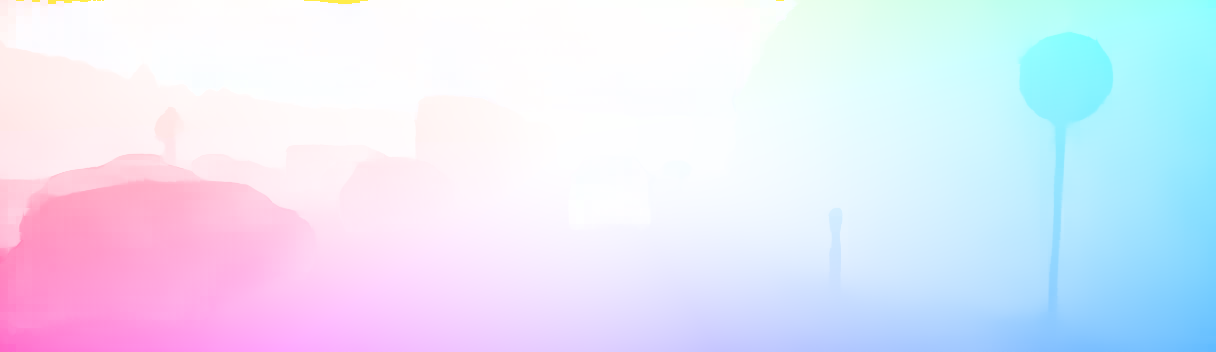}}
    \\    
    \caption{\textbf{Examples of DRAFT predictions} on the KITTI dataset. From (a) input RGB images, we show (b) optical flow estimates, (c) depth estimates, and (d) projected optical flow from depth and scene flow estimates.}
    \label{fig:qualitative}
    \vspace{-3mm}
\end{figure*}

%% file: figures/comparison.tex
\begin{figure*}[h!]
\vspace{-2mm}
    \centering
    \scriptsize
    \rotatebox{90}{\hspace{4mm} RGB}
    \subfloat{
        \includegraphics[width=0.3\textwidth,height=1.4cm]{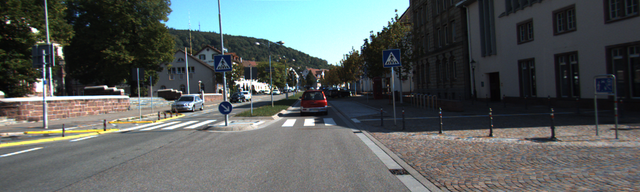}
    }
    \subfloat{
        \includegraphics[width=0.3\textwidth,height=1.4cm]{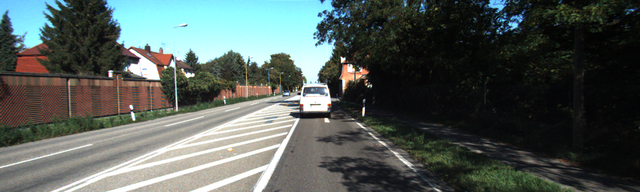}
    }
    \subfloat{
        \includegraphics[width=0.3\textwidth,height=1.4cm]{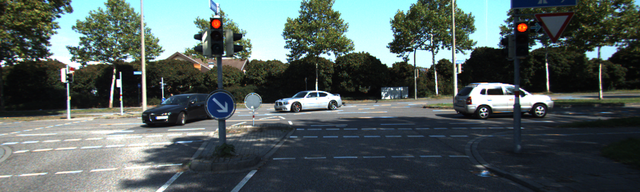}
    }
    \\
    \vspace{-3mm}
    \rotatebox{90}{\hspace{-1mm} ManyDepth}
    \subfloat{
        \includegraphics[width=0.3\textwidth,height=1.4cm]{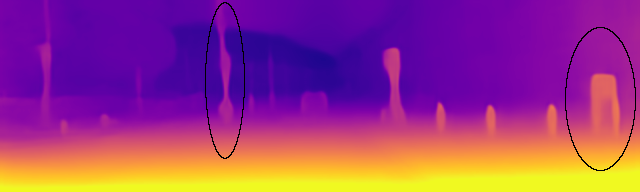}
    }
    \subfloat{
        \includegraphics[width=0.3\textwidth,height=1.4cm]{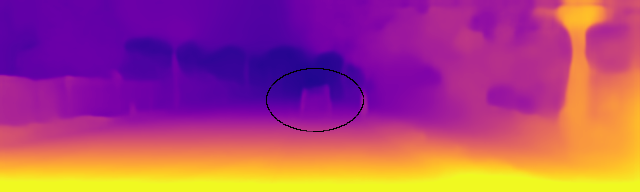}
    }
    \subfloat{
        \includegraphics[width=0.3\textwidth,height=1.4cm]{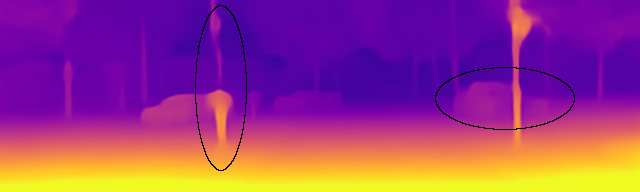}
    }
    \\
    \vspace{-3mm}
    \rotatebox{90}{\hspace{1mm} DRAFT}
    \setcounter{subfigure}{0}
    \subfloat{
        \includegraphics[width=0.3\textwidth,height=1.4cm]{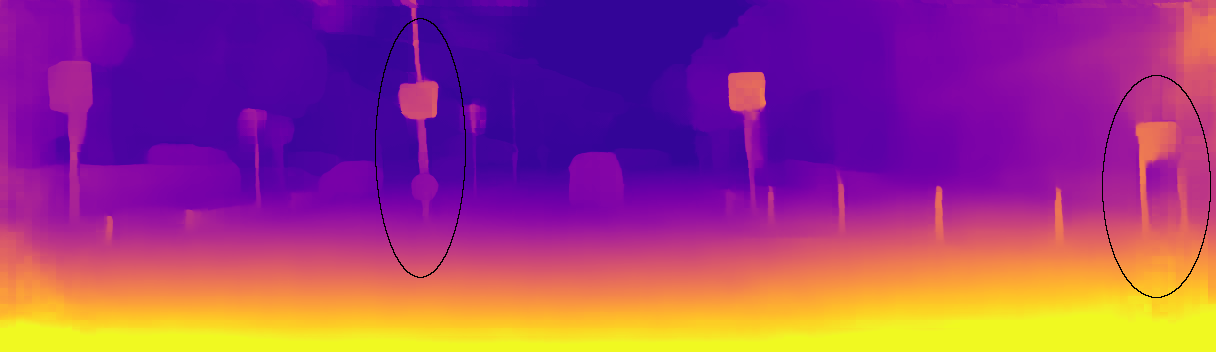}
    }
    \subfloat{
        \includegraphics[width=0.3\textwidth,height=1.4cm]{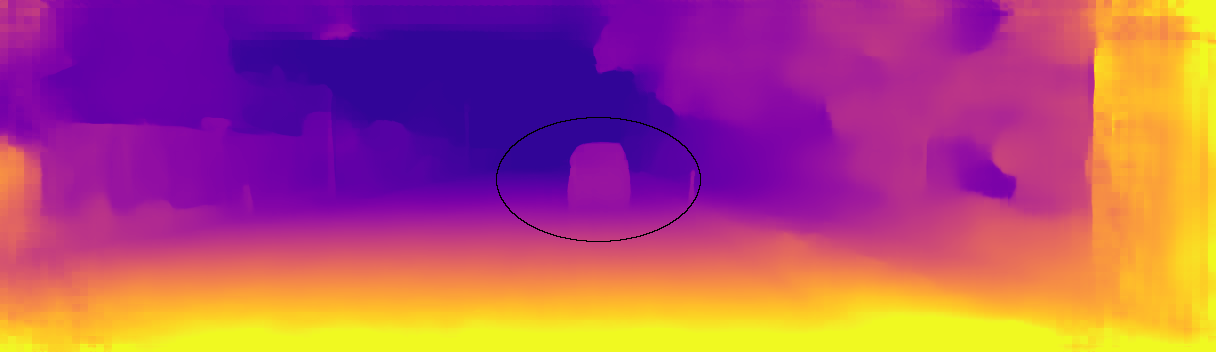}
    }
    \subfloat{
        \includegraphics[width=0.3\textwidth,height=1.4cm]{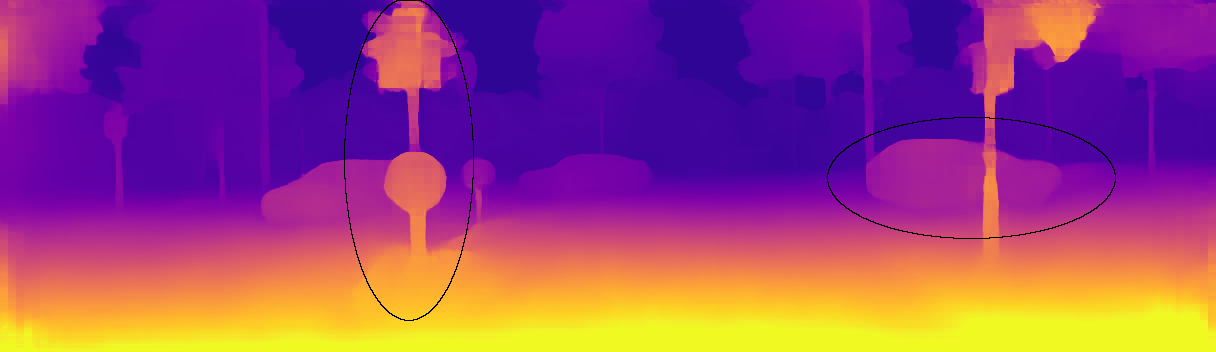}
    }
    \caption{\textbf{Qualitative results} from DRAFT and ManyDepth~\cite{manydepth}. Our proposed architecture produces sharper boundaries and better modeling of dynamic objects.}
    \label{fig:kitti_comparison}
    \vspace{-5mm}
\end{figure*}

%% file: sections/05conclusion.tex
\section{Conclusion}

This paper introduces DRAFT, a novel multi-task deep learning architecture for the joint estimation of optical flow, depth, and scene flow in a self-supervised monocular setting. We connect these tasks in two ways: (i) at the feature level by leveraging optical flow predictions as initialization for depth and scene flow refinement, and (ii) via a series of consistency losses designed to enforce geometric constraints as part of the optimization process.  We use two main sources of supervision: synthetic data combined with real-world self-supervision on unlabeled videos. DRAFT simultaneously achieves state-of-the-art results in optical flow, depth and scene flow on the KITTI dataset, outperforming other methods that focus on a single task.